\documentclass{article}

\usepackage{arxiv}

\usepackage[utf8]{inputenc} 
\usepackage[T1]{fontenc}    
\usepackage{hyperref}       
\usepackage{url}            
\usepackage{booktabs}       
\usepackage{amsmath}
\usepackage{amsfonts}       
\usepackage{nicefrac}       
\usepackage{microtype}      
\usepackage{lipsum}
\usepackage{graphicx}
\usepackage{caption}
\usepackage{subcaption}
\usepackage{makecell}
\usepackage{authblk}
\graphicspath{ {./images/} }

\title{Training an MoE LLM in MXFP4: How We Closed the Gap to BF16}

\author{Han Wang}
\author{Zhitao Wang}
\author{Natalia Frumkin}
\author{Xuanwu Yin}
\author{Alireza Khodamoradi}
\author{Dong Li}
\author{Emad Barsoum}
\affil{Advanced Micro Devices, Inc. \\ \{Han.Wang, Zhitao.Wang, Natalia.Frumkin, Xuanwu.Yin, alireza.khodamoradi, d.li, Emad.Barsoum\}@amd.com}
\date{}

\providecommand{\tightlist}{%
  \setlength{\itemsep}{0pt}\setlength{\parskip}{0pt}}

\begin{document}
\maketitle
\begin{abstract}
Training large language models (LLMs) in sub-8-bit arithmetic promises substantial gains in memory bandwidth and compute throughput, yet FP4 formats introduce quantization error severe enough to destabilize optimization if left unaddressed. We present \href{https://github.com/AMD-AGI/ALTO}{ALTO}, an open-source training recipe for \textbf{GPT-OSS-20B}, a 20-billion-parameter mixture-of-experts (MoE) model, on the \textbf{MLPerf Small MoE Training} benchmark under the \textbf{MXFP4} microscaling format. The recipe combines established techniques from Abecassis, Felix et al.~\cite{abecassis2025pretraining} (hybrid 1D/2D block quantization, randomized Hadamard transforms, and stochastic rounding) with a simplified weight de-oscillation scheme adapted from TetraJet-v2~\cite{chen2025tetrajet}. On C4 validation at global batch size 16, \textbf{MXFP4 + RHT + SR + de-oscillation} reaches a validation loss of 3.3350 at 16,128 steps, only 0.007 above the BF16 baseline (3.3283), closing roughly \textbf{50\%} of the gap left by MXFP4 without de-oscillation (3.3418). Measured by steps to reach the MLPerf quality target (validation loss 3.34), de-oscillation cuts the FP4 convergence overhead relative to BF16 from \textbf{+20.0\%} to \textbf{+5.0\%}. We also report negative end-to-end results from differential gradient estimation, outlier clipping, and macro-block scaling, despite each showing operator-level SNR gains. All experiments use simulated MXFP4 kernels on AMD MI300X hardware and wall-clock speed has not been optimized.
\end{abstract}

\section{Introduction}\label{introduction}
Low-precision training means representing the weights, activations, and gradients used during pretraining in a narrower numerical format than the FP32/BF16 baseline, in exchange for some loss of numerical fidelity. This trade-off matters because operand size directly bounds two of the main costs of training large models: memory bandwidth and compute throughput. Fewer bits per value means less data to move between memory and compute units and, on hardware with matching low-precision matrix units, more throughput per GEMM. This is why the field has kept pushing bit width down, from FP16 and BF16 to FP8 and, more recently, to 4-bit floating-point formats standardized under the Open Compute Project (OCP) Microscaling (MX) specification~\cite{ocpmxfpspec}. MXFP4 encodes weights and activations as E2M1 (signed, 2 \textbf{E}xponent, and 1 \textbf{M}antissa bit) values with per-block UE8M0 (\textbf{U}nsigned, 8 \textbf{E}xponent, and 0 \textbf{M}antissa bit) scales, yielding a theoretical $4\times$ reduction in operand footprint relative to BF16. For sparse MoE models such as GPT-OSS-20B~\cite{openai2025gptoss120bgptoss20bmodel}, where expert matrices dominate compute and memory, MXFP4 training could unlock disproportionate efficiency gains, since the largest tensors in the model are exactly the ones the format targets.

Realizing this in practice is not merely a matter of substituting GEMM kernels, because FP4 quantization interacts with training dynamics in ways that milder formats do not. One phenomenon central to this report is weight \emph{oscillation}: the optimizer updates a full-precision master weight, but the GEMM only ever sees its quantized value, and that quantized value can jump between representable levels from step to step even when the underlying master weight is moving smoothly and by a negligible amount. This happens for two related reasons. First, an element sitting near a quantization bin boundary can flip which bin it rounds into from a tiny update, so the quantized weight oscillates between adjacent lower and higher bins while the full-precision weight barely moves. Second, an outlier element within a block can dominate that block's shared scale, so as gradients and neighboring values fluctuate, the scale shifts enough to repeatedly push the quantized weight across bin boundaries, even when the element itself is not near a boundary. Beyond oscillation, FP4 quantization more generally amplifies sensitivity to tensor outliers, and straight-through estimation (STE) of gradients through non-differentiable quantizers introduces its own bias. All of these effects compound over long-horizon pretraining runs, and oscillation in particular tends to become more damaging later in training, once the loss enters a fine-grained convergence regime.

This paper describes our methodology for closing the accuracy gap between MXFP4 and BF16 training on the MLPerf Small MoE training task \cite{mlperfsmallmoe}. 2D block quantization, randomized Hadamard transforms, and stochastic rounding are adopted from Abecassis, Felix et al.~\cite{abecassis2025pretraining} (Section~\ref{hybrid-1d2d-block-quantization}~--~\ref{stochastic-rounding-on-gradients}). Our focus is on composing these methods into a training recipe suited to MX-format MoE models, implementing them in a unified stack based on Torchtitan~\cite{liang2025torchtitan}, and evaluating which combinations matter at GPT-OSS-20B scale. Weight de-oscillation is adapted from TetraJet-v2~\cite{chen2025tetrajet}. We additionally compare it against low-rank outlier compensation as a design choice. We state explicitly where the implementation remains a research prototype rather than a performance-optimized production
system.

\section{Related Work}\label{related-work}

\paragraph{Low-precision training.} Reduced-precision arithmetic has been a primary lever for scaling neural-network training. Mixed-precision FP16 training with loss scaling~\cite{micikevicius2018mixed} and 8-bit floating-point training~\cite{wang2018training} established that sub-FP32 formats can preserve convergence. The standardization of FP8
formats (E4M3/E5M2) for deep learning~\cite{micikevicius2022fp8} and the FP8 pretraining of
DeepSeek-V3~\cite{deepseekai2024deepseekv3} demonstrated that 8-bit
formats are viable for frontier-scale LLMs. Our work extends this
trajectory to 4-bit floating point, where quantization error is
substantially larger and additional stabilization is required.

\paragraph{Microscaling and FP4 training.} The OCP Microscaling (MX)
specification~\cite{ocpmxfpspec} and the underlying study of microscaling
data formats~\cite{rouhani2023microscaling} define block-scaled low-bit
formats (MXFP8, MXFP6, MXFP4) that share a per-block scale to preserve
dynamic range. Recent efforts push training into FP4: Abecassis, Felix's NVFP4
pretraining recipe~\cite{abecassis2025pretraining} introduces hybrid
block quantization, randomized Hadamard transforms, and stochastic
rounding, which we adopt and adapt. Wang, Ruizhe et al.~\cite{wang2025optimizing} optimizes
FP4 training with a differentiable gradient estimator and outlier
handling. Tseng, Albert et al.~\cite{tseng2025training} studies MXFP4 training with stochastic
rounding and Hadamard-based error reduction. And Chhugani, Jatin et al.~\cite{chhugani2026unveiling} propose macro-block scaling to reduce MXFP4
quantization error. TetraJet-v2~\cite{chen2025tetrajet} targets accurate
NVFP4 training through oscillation suppression and outlier control, and
is the basis for our simplified weight de-oscillation. Our contribution
is the composition and empirical evaluation of these techniques on an
MoE model at GPT-OSS-20B scale.

\paragraph{Outlier mitigation and rotations.} Activation and weight
outliers are the dominant obstacle to aggressive quantization. Prior
work isolates outlier channels in mixed precision (LLM.int8()~\cite{dettmers2022llmint8}), performs error-compensated post-training
quantization (GPTQ~\cite{frantar2023gptq}), or removes outliers via
rotation. QuaRot~\cite{ashkboos2024quarot} and SpinQuant~\cite{liu2024spinquant} apply rotations, including Hadamard
transforms, to produce outlier-free representations amenable to 4-bit
inference. QuEST~\cite{panferov2025quest} pushes stable training toward
very low bit widths. The randomized Hadamard transform and outlier
clipping in our recipe follow this rotation- and clipping-based line of
work, applied on the training path rather than post hoc.

\paragraph{Weight oscillation in quantized training.} Quantization-aware
training is prone to weight oscillation, where quantized weights flip
between adjacent bins despite smooth full-precision trajectories. Nagel, Markus et al.~\cite{nagel2022overcoming} first analyzed and mitigated this phenomenon
in QAT, and TetraJet-v2~\cite{chen2025tetrajet} revisits it for FP4
pretraining via its OsciReset mechanism. Our weight de-oscillation
(Section~\ref{weight-de-oscillation}) is a simplified variant tailored to
MX-format MoE training.

\paragraph{Mixture-of-experts models.} Sparsely activated MoE
architectures scale model capacity while keeping per-token compute
bounded~\cite{shazeer2017outrageously,lepikhin2020gshard,fedus2022switch}.
GPT-OSS-20B~\cite{openai2025gptoss120bgptoss20bmodel} is a recent
open-weight MoE model and the target of the MLPerf Small MoE training
benchmark~\cite{mlperfsmallmoe}. Because expert matrices dominate compute
and memory in such models, they are natural beneficiaries of FP4
training, which motivates our focus on this setting.

\section{Problem Setting}\label{problem-setting}

\subsection{Benchmark: MLPerf Small MoE Training}\label{benchmark-mlperf-small-moe}

We evaluate on the MLPerf Small MoE training benchmark~\cite{mlperfsmallmoe}, which specifies pretraining of a sparse MoE LLM from scratch and measuring convergence to a fixed validation-loss target. Our target model is GPT-OSS-20B~\cite{openai2025gptoss120bgptoss20bmodel}, an open-source 20-billion-parameter mixture-of-experts architecture released by OpenAI.

\textbf{Dataset.} Training and evaluation use the C4 dataset~\cite{raffel2020exploring}, which is pre-tokenized with the Llama-3.1 tokenizer~\cite{grattafiori2024llama3herdmodels}. Training reads the
\texttt{c4-train.en\_6\_text\_document} shards; validation uses the
\texttt{c4-validation-91205-samples.en\_text\_document} split. The
preprocessed dataset is roughly 80 GB.

\textbf{Quality target.} The benchmark's convergence
criterion is a \textbf{validation loss (log perplexity) of 3.34}.
Validation is performed every 12,288 samples (768 iterations at GBS=16)
over the first 1,024 samples of the validation set.

\subsection{Training Protocol}\label{training-protocol}

We adopt the same task, dataset, target loss, and evaluation cadence as the MLPerf Small MoE training benchmark~\cite{mlperfsmallmoe} and incorporate MXFP4/NVFP4 linear/grouped-GEMM operators from our \textbf{A}dvanced \textbf{L}ow-precision \textbf{T}raining and \textbf{O}ptimization (ALTO) project~\cite{alto}. Training hyperparameters follow the default configuration in Table~\ref{tab:hyperparam-gptoss20b}.

\begin{table}
 \caption{Hyperparameters of GPT-OSS-20B pretrain.}
  \centering
  \begin{tabular}{ll}
    \toprule
    Hyperparameter & Value                   \\
    \midrule
    Sequence length & 8,192 tokens \\
    Global batch size & 16 \\
    Max training steps & 1,200,000 \\
    Base learning rate & \(4 \times 10^{-4}\) \\
    Min learning rate & \(4 \times 10^{-5}\) \\
    LR scheduler & cosine decay, 128-step warmup \\
    Optimizer & AdamW (\(\beta_1=0.9\), \(\beta_2=0.95\), weight decay
    \(0.1\)) \\
    Parallelism & Expert parallelism degree 8 \\
    \bottomrule
  \end{tabular}
  \label{tab:hyperparam-gptoss20b}
\end{table}

MXFP4 is applied to all \texttt{Linear} layers and
\texttt{GroupedExperts} modules. The \texttt{lm\_head} projection and
MoE router gates are excluded from quantization, as routing decisions
are sensitive to small perturbations.

\subsection{Evaluation Methodology}\label{evaluation-methodology}

We evaluate at two levels: operator-level numerical accuracy of
individual MXFP4 kernels, and end-to-end validation loss during
full-model pretraining.

\textbf{Synthetic data with injected outliers.} Operator-level accuracy tests use a
fixed synthetic generator designed to stress block-wise quantization
under sparse heavy tails rather than i.i.d. Gaussian noise alone. Each tensor is drawn from the following distribution:

\[
\mathcal{N}(0, 1) + \mathrm{Bernoulli}(0.005) \odot \mathcal{N}(0, 10000),
\]

Each element independently receives an additional outlier
perturbation with probability \(0.5\%\), scaled to roughly $100\times$ standard
deviation. Inputs, weights, and loss targets are all generated by this
procedure.

\textbf{Operator-level metric.} Before scaling to end-to-end runs, we
verify that ALTO's MXFP4 Linear and grouped-GEMM autograd paths
reproduce a high-precision reference. Each configuration is evaluated by
running a forward pass and a full backward pass through an MSE loss,
then comparing the MXFP4 outputs and gradients against a BF16 reference
that performs the same computation without MXFP4 quantization. We report
signal-to-noise ratio (SNR) in decibels:

\[
\mathrm{SNR} = 10 \log_{10} \frac{\sum_i \mathbf{X}_i^2}{\sum_i (\mathbf{X}_i - \mathbf{\hat{X}}_i)^2},
\]

where \(\mathbf{X}\) is the BF16 reference tensor and
\(\mathbf{\hat{X}}\) is the MXFP4 result. SNR is computed in FP32
accumulation for numerical stability. We report three quantities per
configuration: forward output (\(\mathbf{O}\)), input gradient
(\(\mathrm{d}\mathbf{X}\)), and weight gradient
(\(\mathrm{d}\mathbf{W}\)).

\textbf{Steps-to-target metric.} In addition to absolute validation
loss, we report a convergence-speed metric: the number of training steps
required to first reach a validation loss \(\leq 3.34\), the MLPerf
Small MoE quality target (Section~\ref{benchmark-mlperf-small-moe}). Because validation is only
recorded at fixed checkpoints (768-step cadence at GBS=16), we report
the earliest checkpoint at which the target is met. This metric captures
how quickly each low-precision recipe attains the reference quality bar,
complementing the final-loss comparison.

\section{Method}\label{method}

We organize accuracy recovery methods into a composable stack. The
recommended recipe enables hybrid block quantization, Randomized Hadamard Transform (RHT), and Stochastic Rounding (SR). Weight de-oscillation is applied in the late-training stage.

\begin{figure}
    \centering
    \includegraphics[width=\textwidth]{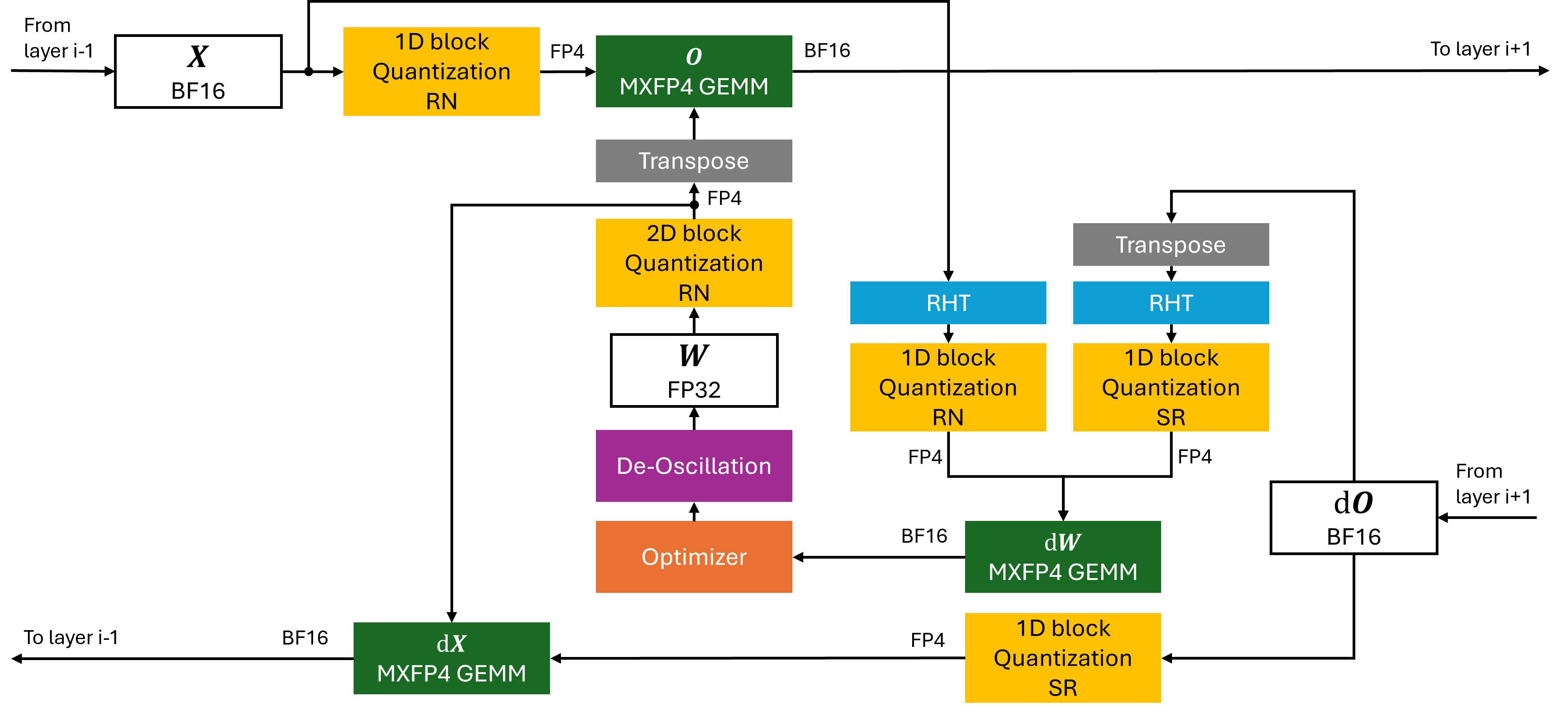}
    \caption{Compute flow of an MXFP4 linear layer.}
    \label{fig:flow}
\end{figure}

\textbf{MXFP4 training paradigm.} MXFP4 is a block-scaled 4-bit format
under the OCP Microscaling (MX) specification: each block of 32
consecutive elements shares a UE8M0 scale factor \(s\), with elements
quantized to E2M1 (approximate range \([-6, 6]\)). Given a block \(\mathbf{x}\),

\[
s = \frac{\max_i |x_i|}{r_{\max}}, \qquad
\hat{x}_i = \mathcal{Q}_{\mathrm{E2M1}} \left(\frac{x_i}{s}\right), \qquad
\tilde{x}_i = \hat{x}_i \cdot s,
\]

where \(\mathcal{Q}_{\mathrm{E2M1}}\) rounds to the nearest
representable E2M1 value and \(r_{\max}\) is the largest E2M1 magnitude.
As illustrated in Figure~\ref{fig:flow}, each linear layer has three underlying
GEMMs: a GEMM in the forward pass producing layer output \(\mathbf{O}\),
and separate GEMMs producing activation gradient
(\(\mathrm{d}\mathbf{X}\)) and weight gradient
(\(\mathrm{d}\mathbf{W}\)) in the backward pass. GEMM operations consume
FP4 tensors as inputs and produce outputs in BF16 or FP32. Gradients
flow through the quantizer via straight-through estimation (STE) unless
modulated by the differential gradient estimation (DGE). On
hardware without native MXFP4 GEMM support, ALTO performs \textbf{simulated
quantization}, which means operands are quantized to MXFP4 and immediately
dequantized back to BF16/FP32 before a standard high-precision
GEMM, faithfully reproducing FP4 numerical error without realizing
memory-bandwidth savings (Section~\ref{limitations}).

Sections~\ref{hybrid-1d2d-block-quantization}~--~\ref{stochastic-rounding-on-gradients} describe techniques adopted from Abecassis, Felix et al.~\cite{abecassis2025pretraining} and integrated in ALTO. Section~\ref{weight-de-oscillation} introduces our weight de-oscillation method, adapted from the OsciReset algorithm in TetraJet-v2~\cite{chen2025tetrajet}. Sections~\ref{differential-gradient-estimation-dge}~--~\ref{low-rank-outlier-compensation} cover additional methods explored but not retained in the recommended recipe.

\subsection{Hybrid 1D/2D Block Quantization}\label{hybrid-1d2d-block-quantization}

The MX specification defines block scaling along
contiguous 1D blocks only: each UE8M0 scale is shared by 32 consecutive
elements along a single axis. For inference, quantizing each operand
once along its contraction axis suffices; training imposes additional constraints.
Consider a linear layer with weight
\(\mathbf{W} \in \mathbb{R}^{N \times K}\) and forward pass
\(\mathbf{O} = \mathbf{X}\cdot\mathbf{W}^{\top}\). The forward GEMM
contracts along \(K\), so \(\mathbf{W}\) is naturally quantized with
scales aligned to that axis. The backward pass
\(\mathrm{d} \mathbf{X} = \mathrm{d}\mathbf{O}\cdot\mathbf{W}\) accesses
\(\mathbf{W}\) along \(N\).
Under 1D MX blocking, forward and backward therefore require quantization along different axes. A strictly 1D weight layout necessitates two separate quantizations, doubling overhead while still leaving the forward and backward views inconsistent.
Activations present a separate concern: outlier redistribution via Randomized Hadamard Transform (RHT, Section~\ref{randomized-hadamard-transform}) operates on 1D segments and is
incompatible with 2D activation blocking.

To reconcile these requirements, we adopt a \textbf{1D--2D hybrid} layout (denoted
\emph{1d2d}) that respects the MX spec for activations while extending
it for weights, following Abecassis, Felix et al.~\cite{abecassis2025pretraining}:

\begin{itemize}
\tightlist
\item
  \textbf{Activations --- Quantized in 1D block (32-element block; canonical MX). } Compatible with RHT on the 1D activation path (Section~\ref{randomized-hadamard-transform}) and lower quantization error.
\item
  \textbf{Weights --- Quantized in 2D block ($32 \times 32$ block; beyond MX spec).} Quantization valid for forward and backward, lower quantization overhead, and eliminating forward--backward weight discrepancy.
\end{itemize}

2D block weight quantization partitions \(\mathbf{W}\) into $32 \times 32$
blocks, each sharing one scale. Because the block grid spans both
spatial dimensions, the same quantized representation is valid whether
\(\mathbf{W}\) is consumed in the forward orientation or transposed in
the backward pass. As a secondary benefit, 2D blocking reduces
quantization overhead: single quantization step for both forward and
backward passes, and reduced memory footprint for scale metadata. We
adopt $32 \times 32$ blocks for MXFP4 format and $16\times16$ for NVFP4.

For MoE grouped GEMM, expert weights
\(\mathbf{W} \in \mathbb{R}^{E \times N \times K}\) receive 2D blocking
along the final two dimensions. One quantized expert weight tensor is
reused across both the forward and the backward kernels, with no
axis-dependent re-quantization and no discrepancy between the weight
operand presented to the forward and backward graphs.

\subsection{Randomized Hadamard Transform}\label{randomized-hadamard-transform}

A small number of large-magnitude activations
(outliers) inflate the per-block scale, compressing the remaining
elements and degrading signal fidelity. Following Abecassis, Felix et al.~\cite{abecassis2025pretraining}, we apply a
\textbf{randomized Hadamard transform} (RHT) on the weight gradient
path to redistribute outliers before quantization. Let \(\mathbf{H} \in \mathbb{R}^{32 \times 32}\) be a Hadamard matrix constructed via Sylvester's recursion, further randomized by
random sign flips and column permutations. In the weight-gradient GEMM,
activations are transformed as \(\mathbf{X}' \leftarrow \mathbf{H}\mathbf{X}\);
the output gradient is transformed similarly \(\mathrm{d}\mathbf{O}' \leftarrow \mathbf{H}\mathrm{d}\mathbf{O}\).
Because \(\mathbf{H}\) is orthogonal (up to scaling), the exact weight
gradient satisfies

\[
\mathrm{d}\mathbf{W} = \left(\mathbf{H}\mathrm{d}\mathbf{O}\right)^{\top} \left(\mathbf{H}\mathbf{X}\right) = \mathrm{d}\mathbf{O}^{\top}\mathbf{H}^{\top}\mathbf{H}\mathbf{X} = \mathrm{d}\mathbf{O}^{\top}\mathbf{X}
\]

Because \(\mathbf{H}\) is orthogonal (up to scaling), the transform
redistributes energy across coordinates without altering the underlying
linear operator's gradient in expectation over randomization. RHT is
applied exclusively on the 1D activation path and is therefore mutually
exclusive with 2D activation blocking, motivating the hybrid layout of
Section~\ref{hybrid-1d2d-block-quantization}.

\subsection{Stochastic Rounding on Gradients}\label{stochastic-rounding-on-gradients}

Deterministic round-to-nearest-even quantization of
gradients introduces a systematic bias: small gradient components are
disproportionately rounded to zero, attenuating effective learning rates
in FP4.
As in Abecassis, Felix et al.~\cite{abecassis2025pretraining}, we employ
\textbf{stochastic rounding} (SR) during gradient quantization in the
backward pass to mitigate this effect. Each element is rounded up or down with probability
proportional to its fractional distance from the two adjacent
representable values, yielding an unbiased estimator of the
pre-quantization gradient in expectation. On CDNA4, stochastic rounding
is accelerated through inline assembly
(\texttt{v\_cvt\_scalef32\_sr\_pk\_fp4\_*}). Forward-pass quantization
of weights and activations remains deterministic.

\subsection{Weight De-Oscillation}\label{weight-de-oscillation}

Weight oscillation arises from two complementary
mechanisms. First, when a master weight \(w\) resides near a
quantization bin boundary, infinitesimal AdamW updates can cause
\(Q(w)\) to alternate between adjacent bins while the FP trajectory
remains smooth. Second, outlier elements with disproportionately
large magnitude within a block can drive the block scale and
repeatedly push \(Q(w)\) across bin boundaries as gradients and
neighboring values fluctuate, even when \(w\) itself does not sit at a
boundary. In both cases, the weight seen by the GEMM oscillates, a
pathology analyzed in TetraJet-v2~\cite{chen2025tetrajet}.

\begin{figure}
    \centering
    \includegraphics[width=\textwidth]{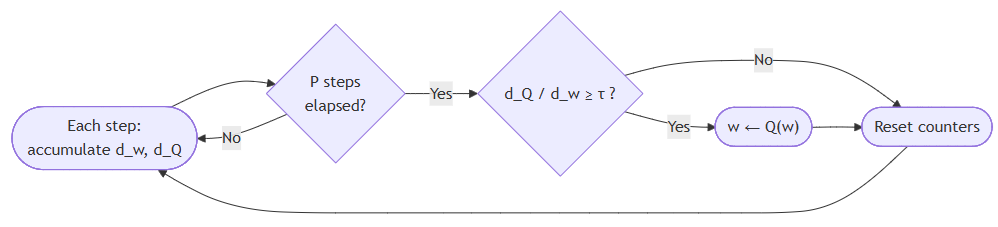}
    \caption{Weight de-oscillation core loop (post--AdamW step hook; period \(P\), threshold \(\tau\)).}
    \label{fig:deosc}
\end{figure}

We implement a weight de-oscillation method adapted from TetraJet-v2~\cite{chen2025tetrajet}. As shown in Figure~\ref{fig:deosc}, over a window of \(P\) optimizer steps (default \(P=200\)), we accumulate per-element L1 travel distances:

\[
d_w = \sum_{t=1}^{P} |w_t - w_{t-1}|, \qquad
d_{Q} = \sum_{t=1}^{P} |Q(w_t) - Q(w_{t-1})|.
\]

An element is flagged as oscillating when the distance ratio
\(\mathrm{DistRatio} = d_{Q} / d_w\) exceeds a threshold \(\tau\)
(default \(\tau=4.0\)). This criterion captures both bin-boundary
flicker and outlier-induced scale sensitivity, since either mechanism
produces large quantized movement relative to small FP movement. At the
end of each window, flagged elements are snapped to their current bin
center: \(w \leftarrow Q(w)\). The procedure is installed as a post-step
hook on AdamW and is activated after a warmup of \texttt{deosc\_step}
training steps (default 2,000 for GBS=16).

Our implementation shares the same oscillation criterion as TetraJet-v2~\cite{chen2025tetrajet}, the distance ratio $\mathrm{DistRatio} = d_Q/d_w$, but differs from TetraJet-v2 in two respects:

\begin{itemize}
\tightlist
\item
  \textbf{Model-decoupled, optimizer-level integration.} TetraJet-v2~\cite{chen2025tetrajet} is embedded in the model definition: it requires model-side changes that expose per-layer quantized weights and inject the reset logic into the forward/backward path. ALTO instead implements de-oscillation entirely at the optimizer level, as a post-step hook on AdamW that operates on the parameters and their MXFP4/NVFP4 wrappers. Because the mechanism reads only the local copy of master weights through the optimizer, it is agnostic to model architecture and requires no changes to the model architecture, making it reusable across models.
\item
  \textbf{Different hyperparameters.} We use an earlier activation step and a lower flag threshold than TetraJet-v2~\cite{chen2025tetrajet}. De-oscillation is enabled after a warmup of \texttt{deosc\_step} steps (default 2,000 at GBS=16), whereas TetraJet-v2 starts at step 8,000. We also flag oscillating elements at $\tau=4.0$ versus TetraJet-v2's $\tau=16$, a lower threshold that treats oscillation more aggressively. Both settings are tuned to our MXFP4 training horizon.
\end{itemize}

\subsection{Differential Gradient Estimation (DGE)}\label{differential-gradient-estimation-dge}

The straight-through estimator (STE) back-propagates
through FP4 weight quantization as if the quantizer were the identity,
discarding all information about local sensitivity within each
representable bin. This yields a biased gradient that misrepresents how
a weight perturbation actually moves the quantized value. Wang, Ruizhe et al.~\cite{wang2025optimizing} address this with a \textbf{differentiable
gradient estimator} (DGE) that approximates the gradient of the
quantization mapping with a smooth function whose magnitude reflects
local sensitivity within each bin. But their estimator is not continuous
across bin boundaries, introducing discontinuities in the gradient
field.

\begin{figure}
    \centering
    \begin{subfigure}[b]{0.5\textwidth}
        \centering
        \includegraphics[width=0.8\textwidth]{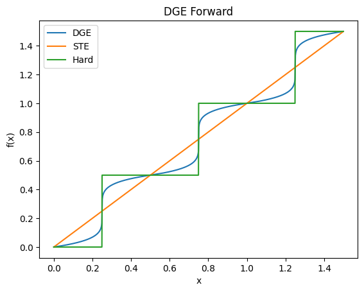}
        \caption{Forward function.}
    \end{subfigure}%
    ~ 
    \begin{subfigure}[b]{0.5\textwidth}
        \centering
        \includegraphics[width=0.8\textwidth]{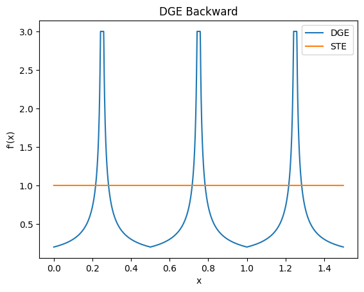}
        \caption{Backward function.}
    \end{subfigure}
    \caption{Comparison between Differential Gradient Estimation (DGE) and Straight-Through Estimation (STE).}
    \label{fig:dge}
\end{figure}

We adopt the DGE concept but use a corrected formula:
ALTO employs a piecewise power-law surrogate that is continuous between
segments (as shown in Figure~\ref{fig:dge}). For a value \(x\) lying in a bin of width \(\delta\) with midpoint \(m\), and smoothing parameter \(k=5\), we define

\[
f(x) = \left(\frac{\delta}{2}\right)^{1 - 1/k}\cdot\mathrm{sgn}(x - m)\cdot|x - m|^{1/k} + m,
\]

\[
f'(x) = \frac{1}{k}\left(\frac{\delta}{2}\right)^{1 - 1/k}\cdot|x - m|^{1/k - 1},
\]

where \(f'(x)\) is clamped to \([0, 3]\) prior to application. The weight gradient is scaled as

\[
\mathrm{d}\mathbf{W} \leftarrow \mathrm{d}\mathbf{W} \odot f'(\mathbf{W}).
\]

For MXFP4, bin widths \(\delta\) and midpoints \(m\) are determined by the E2M1 representable grid.

\subsection{Outlier Clipping}\label{outlier-clipping}

A few large-magnitude elements inflate the per-block
scale, consuming representable range and compressing or saturating the
remaining elements, which lowers effective precision for the bulk of the
distribution. We cap outliers before quantization to reclaim
precision for the majority of elements. Two clipping modes are
supported:

\begin{itemize}
\tightlist
\item
  \textbf{Static clipping} scales inputs by \(3/4\) before quantization
  and by \(4/3\) after dequantization, with a compensating factor of
  \(16/9\) on the weight-gradient path to preserve gradient consistency.
\item
  \textbf{Dynamic clipping} follows QuEST~\cite{panferov2025quest}: 
  a per-block clipping threshold is estimated from the
  block standard deviation,
  \(\hat{m} = (2.922/6)\,\mathrm{std}(\mathbf{x}_{\mathrm{block}})\),
  and requires co-use with RHT in our implementation.
\end{itemize}

\subsection{Macro-Block Scaling}\label{macro-block-scaling}

A single UE8M0 scale shared by each 32-element MX
block is a coarse granularity: when a block contains an outlier, the
shared scale is dominated by that element and the remaining values lose
precision, so outlier-induced saturation is only partially mitigated by
standard MX block scaling. Chhugani, Jatin et al.~\cite{chhugani2026unveiling} propose \textbf{Macro Block Scaling (MBS)} as a two-level scheme for MXFP4: a coarse macro-scale is
applied before the standard 32-element MX block quantization, allocating
higher-precision scaling at a larger granularity to better preserve
outliers. Our configuration applies MBS with $128 \times 128$ blocks on weights
and $1 \times 128$ blocks on activations. Within each macro-block, the scale is
derived so that the block maximum maps to the largest E2M1 representable
magnitude 6.0:

\[
s_{\mathrm{macro}} = \frac{6}{\max_{i \in \mathrm{block}} |x_i|}.
\]

The macro-scale is encoded as a shared mantissa: only the upper 8
mantissa bits of the FP32 scale factor are stored, and the value is
reconstructed for pre-scaling before MXFP4 quantization and descaling
after dequantization.

\subsection{Low-Rank Outlier Compensation}\label{low-rank-outlier-compensation}

FP4 saturation clips the largest-magnitude components
of a layer's computation, discarding information carried by a few
outlier-dominated directions that a single low-precision GEMM cannot
represent. Following the outlier-compensation paradigm, ALTO can
decompose a linear layer into a full MXFP4 GEMM plus a low-rank branch to preserve high magnitude outliers:

\[
\mathbf{O} = \mathbf{X}\mathbf{W}^{\top} + \bigl((\mathbf{X}\mathbf{V}) \odot \boldsymbol{\sigma}\bigr)\mathbf{U}^{\top},
\]

where \(\mathbf{W}, \mathbf{U}, \mathbf{V}\) are quantized to MXFP4 and
\((\mathbf{U}, \mathbf{V}, \boldsymbol{\sigma})\) form a rank-\(r\)
branch intended to capture information lost to FP4 saturation. Each
compensated layer adds two extra MXFP4 parameters (\(\mathbf{U}\),
\(\mathbf{V}\)) with one FP32 parameter (\(\boldsymbol{\sigma}\)) and 6
additional low-rank GEMMs in every forward-backward pass, with cost
scaling with \(r\). We apply this exclusively to linear layers and do not consider MoEs.

\subsection{M+Adam: Multiplicative Update for Adam}
\label{M+Adam}

In low-precision training, Adam~\cite{kingma2017adammethodstochasticoptimization} forms its moment estimates from the quantized gradient \(G^t\), while retaining the optimizer state and update in higher precision:
\begin{equation}
M^t=\beta_1M^{t-1}+(1-\beta_1)G^t,\qquad
V^t=\beta_2V^{t-1}+(1-\beta_2)(G^t)^{\odot 2},
\qquad
U_t^{(a)}=-\eta\frac{\widehat M^t}{\sqrt{\widehat V^t}+\epsilon}.
\end{equation}
Here, \(M^t\) and \(V^t\) are the first- and second-moment estimates at step \(t\), \(\beta_1\) and \(\beta_2\) are their decay rates, and \(\widehat M^t=M^t/(1-\beta_1^t)\) and \(\widehat V^t=V^t/(1-\beta_2^t)\) denote bias-corrected moments. The operator \(\odot\) denotes elementwise multiplication, \(\eta\) is the learning rate, \(\epsilon\) is a numerical-stability constant, and \(U_t^{(a)}\) is the standard additive Adam update.

M+Adam~\cite{liang2026madamlowprecisiontrainingadditivemultiplicative} augments this update with a multiplicative branch. Writing each nonzero weight as \(w=m2^e\), where \(m\) and \(e\) are its mantissa-like and exponent-like coordinates, respectively, the gradient with respect to \(e\) is
\begin{equation}
g_t^{(e)} = \frac{\partial L}{\partial e} (\ln 2),w_tg_t,
\end{equation}
where \(L\) is the training loss, \(w_t\) is the current weight, and \(g_t=\partial L/\partial w_t\) is its gradient. An additional second-moment accumulator normalizes \(g_t^{(e)}\) to produce a relative update \(U_t^{(m)}\). The weights are then updated as
\begin{equation}
W_{t+1} = W_t+W_t\odot U_t^{(m)}+U_t^{(a)},
\end{equation}
where \(W_t\) denotes the weight tensor. The multiplicative term \(W_t\odot U_t^{(m)}\) scales with the current weight magnitude, helping updates cross low-precision quantization intervals, while \(U_t^{(a)}\) remains necessary near zero and for sign changes. This benefit comes at the cost of additional per-weight optimizer state and computation.

\section{Results}\label{results}

\subsection{Operator-Level Numerical Accuracy}\label{operator-level-numerical-accuracy}

Table~\ref{tab:op-accuracy-linear} and Table~\ref{tab:op-accuracy-gmm} report SNR under the operator-level metric of Section~\ref{evaluation-methodology}. The $-\infty$ entries for low-rank compensation are not measurement failures: with the compensation branch enabled, the layer factorizes its parameters across the main MXFP4 weight $\mathbf{W}$ and the low-rank factors $(\mathbf{U}, \mathbf{V}, \boldsymbol{\sigma})$ (Section~\ref{low-rank-outlier-compensation}), so the gradient with respect to $\mathbf{W}$ is structurally different from the weight gradient of the plain BF16 linear used as reference. Forward output $\mathbf{O}$ and input gradient $\mathrm{d}\mathbf{X}$ remain close, since the compensated linear layer exposes the same effective weight as the reference.

\begin{table}
 \caption{SNR (dB) of MXFP4 linear relative to BF16 reference.}
  \centering
  \begin{tabular}{llll}
    \toprule
    Configuration & Forward \(\mathbf{O}\) & Backward \(\mathrm{d}\mathbf{X}\) & Backward \(\mathrm{d}\mathbf{W}\)                 \\
    \midrule
    1d2d & 13.48 & 11.80 & 13.02 \\
    1d2d + RHT & 13.48 & 11.80 & 13.25 \\
    1d2d + RHT + SR \textit{(recommended config)} & 13.48 & 11.02 & 13.04 \\
    1d2d + RHT + Static Clipping & 13.48 & 11.80 & 13.22 \\
    1d2d + RHT + SR + Static Clipping & 13.48 & 11.02 & 13.10 \\
    1d2d + RHT + Dynamic Clipping & 13.48 & 11.80 & 13.23 \\
    1d2d + RHT + SR + Dynamic Clipping & 13.48 & 11.02 & 13.06 \\
    1d2d + RHT + MBS & 14.64 & 12.20 & 13.75 \\
    1d2d + RHT + SR + MBS & 14.64 & 12.20 & 14.04 \\
    1d2d + RHT + Low-rank compensation ($r=32$) & 13.71 & 12.03 & $-\infty$ \\
    1d2d + RHT + SR + Low-rank compensation ($r=32$) & 13.71 & 11.19 & $-\infty$ \\
    \bottomrule
  \end{tabular}
  \label{tab:op-accuracy-linear}
\end{table}

\begin{table}
 \caption{SNR (dB) of MXFP4 grouped GEMM relative to BF16 reference.}
  \centering
  \begin{tabular}{llll}
    \toprule
    Configuration & Forward \(\mathbf{O}\) & Backward \(\mathrm{d}\mathbf{X}\) & Backward \(\mathrm{d}\mathbf{W}\)                 \\
    \midrule
    1d2d & 13.47 & 12.71 & 12.22 \\
    1d2d + RHT & 13.47 & 12.71 & 12.42 \\
    1d2d + RHT + SR \textit{(recommended config)} & 13.47 & 12.26 & 12.33 \\
    1d2d + RHT + Static Clipping & 13.47 & 12.71 & 12.12 \\
    1d2d + RHT + SR + Static Clipping & 13.47 & 12.26 & 12.35 \\
    1d2d + RHT + Dynamic Clipping & 13.47 & 12.71 & 12.42 \\
    1d2d + RHT + SR + Dynamic Clipping & 13.47 & 12.26 & 12.32 \\
    1d2d + RHT + MBS & 14.64 & 14.17 & 14.05 \\
    1d2d + RHT + SR + MBS & 14.64 & 13.76 & 13.72 \\
    \bottomrule
  \end{tabular}
  \label{tab:op-accuracy-gmm}
\end{table}

\paragraph{Discussion.}

\emph{Randomized Hadamard transform.} RHT leaves the forward pass unchanged, since it is applied only to weight gradients in the backward path. It gives a modest, consistent improvement to weight-gradient SNR, consistent with redistributing outliers before 1D activation-gradient quantization.

\emph{Stochastic rounding.} SR lowers input-gradient SNR while leaving forward SNR and weight-gradient SNR essentially unaffected. This is the expected bias-variance trade-off: SR removes systematic rounding bias at the cost of additional quantization noise in the output gradient. In end-to-end training (Section~\ref{end-to-end-training}), this local SNR penalty is outweighed by the bias reduction SR provides.

\emph{Outlier clipping.} Static and dynamic clipping, with or without SR, land close to their corresponding non-clipping rows. Under this synthetic outlier regime, clipping does not materially improve operator accuracy.

\emph{Macro-block scaling (MBS).} MBS yields the largest operator-level gain, improving forward and backward SNR, with the biggest effect on the grouped-GEMM input gradient. This confirms that coarse shared-mantissa pre-scaling (Section~\ref{macro-block-scaling}) helps mitigate outlier-induced saturation before MX block quantization.

\emph{Low-rank outlier compensation.} Adding the rank-32 branch slightly raises forward and input-gradient SNR, consistent with the auxiliary branch preserving a small amount of outliers that a single MXFP4 GEMM saturates. The weight-gradient column is reported as $-\infty$ because, as noted under Table 1, the compensated layer's weight gradient is not the same quantity as a plain linear's and cannot be scored against the reference. The operator-level gains here are modest and, as Section~\ref{end-to-end-training} shows, do not translate into an end-to-end benefit at $r=32$.

\emph{Recommended quantization configuration.} The recommended configuration (1d2d + RHT + SR) keeps forward accuracy essentially at the unmodified baseline while accepting the SR-induced penalty on the input gradient documented above.

\subsection{End-to-End Training}\label{end-to-end-training}

We pretrain GPT-OSS-20B~\cite{openai2025gptoss120bgptoss20bmodel} from scratch on a C4 subset under the protocol
of Section~\ref{training-protocol}, using simulated MXFP4 or NVFP4 GEMMs for all Linear
and grouped-expert layers (except the \texttt{lm\_head} projection and
MoE router gates). Validation cross-entropy is measured as described in
Section~\ref{evaluation-methodology} at fixed training-step checkpoints. We report results at global batch size (GBS) 16 (base learning rate \(4 \times 10^{-4}\), with
de-oscillation enabled at step 2000). All FP4
runs use the 1d2d+RHT+SR recipe unless noted otherwise.

\begin{figure}
    \centering
    \includegraphics[width=0.5\textwidth]{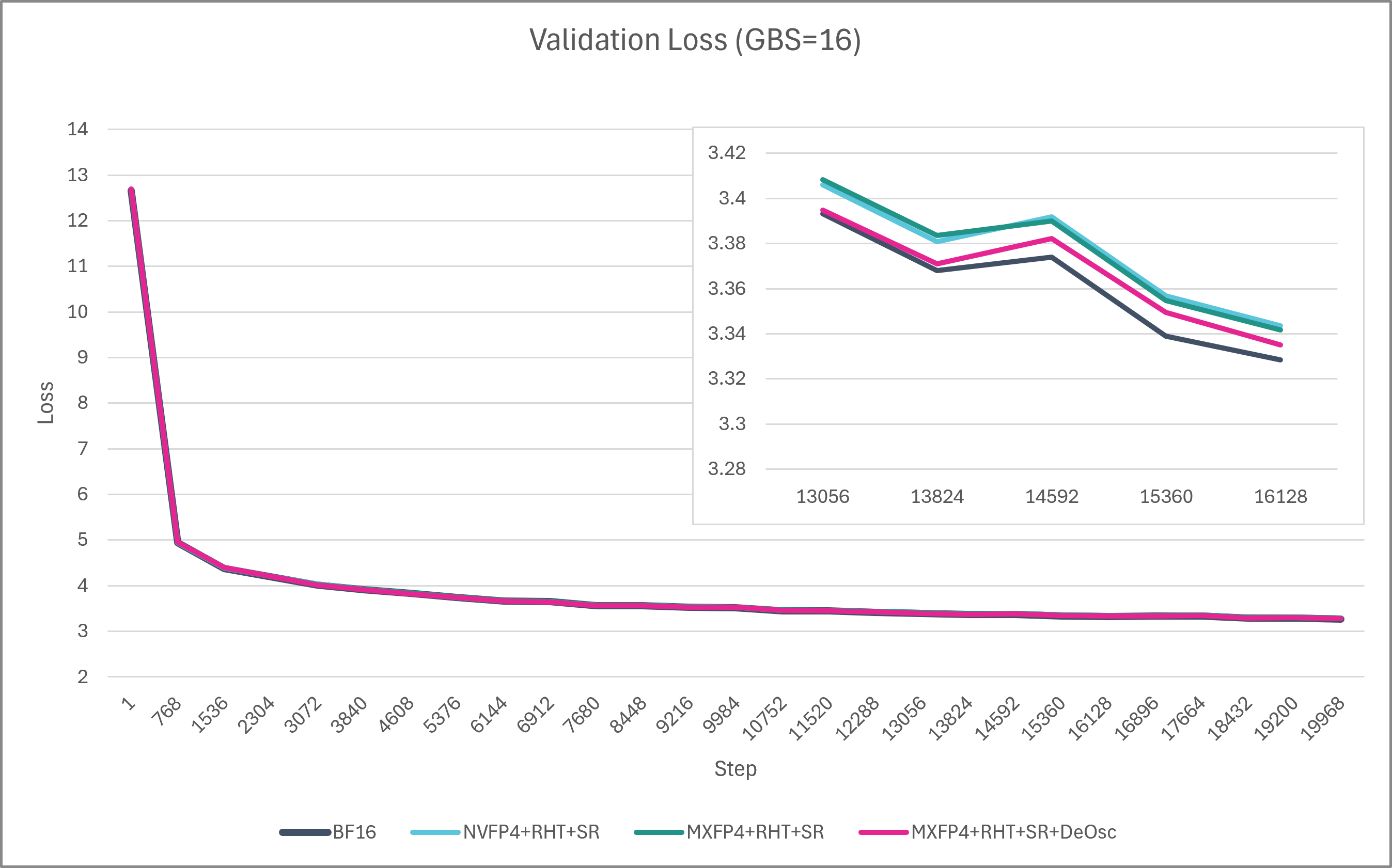}
    \caption{Validation loss on MLPerf Small MoE (GPT-OSS-20B, C4 subset) with GBS=16.}
    \label{fig:valloss16}
\end{figure}

\paragraph{Comparison.} Figure~\ref{fig:valloss16} tracks four training curves through 16,128 steps. The BF16 baseline reaches the lowest validation loss at the final checkpoint. The recommended MXFP4 recipe (1d2d + RHT + SR) trails by a small margin. Adding weight de-oscillation closes roughly half of that remaining gap, and at the final checkpoint MXFP4 with de-oscillation also outperforms out NVFP4 + RHT + SR.

The benefit of de-oscillation shows up in the late-training regime. Earlier in the run, the recipe without de-oscillation is comparable to, or even slightly ahead of, the one with it; de-oscillation only pulls ahead as training approaches the final checkpoint. This pattern is consistent with TetraJet-V2~\cite{chen2025tetrajet}'s design: oscillation accumulates over long horizons and is most harmful once the loss enters a fine-grained convergence phase.

\paragraph{Steps to target validation loss \(3.34\).} Table~\ref{tab:steps-to-target-valloss} reports the
earliest tracked checkpoint at which each recipe crosses the MLPerf
quality target of 3.34 (Section~\ref{evaluation-methodology}). At GBS=16, BF16 reaches the target
first; MXFP4 + RHT + SR + de-oscillation and MXFP4 + RHT + SR + M+Adam follow with only a
\textbf{+5.0\%} step overhead, whereas plain MXFP4 + RHT + SR and NVFP4
+ RHT + SR both incur \textbf{+20.0\%}. De-oscillation and M+Adam hold the FP4
time-to-target overhead to a relatively low percent.

\begin{table}
 \caption{Steps to reach validation loss \(3.34\) (earliest tracked checkpoint).}
  \centering
  \begin{tabular}{lll}
    \toprule
    Method & Steps to target & \(\Delta\) steps vs.~BF16  \\
    \midrule
    BF16 baseline & 15,360 & --- \\
    NVFP4 (1d2d) + RHT + SR & 18,432 & +20.0\% \\
    MXFP4 (1d2d) + RHT + SR & 18,432 & +20.0\% \\
    MXFP4 (1d2d) + RHT + SR + M+Adam & 16,128 & +5.0\% \\
    MXFP4 (1d2d) + RHT + SR + de-oscillation & 16,128 & +5.0\% \\
    \bottomrule
  \end{tabular}
  \label{tab:steps-to-target-valloss}
\end{table}

\paragraph{Ablations.} Table~\ref{tab:ablation-gbs16} isolates individual techniques atop the MXFP4 + RHT + SR base. All ablation rows are drawn from a single sweep in which the shared control (RHT + SR) reaches loss 3.390; $\Delta$ values are computed relative to this control. Several findings follow directly. \textbf{De-oscillation} delivers the largest single improvement ($-0.008$) and achieves the best absolute loss in the ablation. \textbf{MBS} also improves on the base, though by a smaller margin ($-0.004$). \textbf{DGE} and \textbf{dynamic clipping} are strongly detrimental ($+0.201$ and $+0.181$, respectively), far worse than omitting the technique. \textbf{Static clipping} is neutral ($\Delta = 0.000$), while \textbf{low-rank compensation} ($r=32$) is slightly harmful ($+0.007$). \textbf{M+Adam} decomposes each weight into mantissa and exponent components and updates both with coupled Adam-style steps, thereby reducing quantization-induced oscillation at the optimizer level.

\begin{table}
 \caption{Ablation of individual techniques (GBS=16, checkpoint 14,592 steps).}
  \centering
  \begin{tabular}{lll}
    \toprule
    Technique & Validation loss & \(\Delta\) loss vs.~RHT+SR base   \\
    \midrule
BF16 & 3.365/3.371 & -0.025\\
MXFP4 (1d2d) & 3.438 & +0.048 \\
MXFP4 (1d2d) + RHT + SR (base) & 3.390 & —-- \\
+ DGE & 3.591 & +0.201 \\
+ Static clipping & 3.390 & 0.000 \\
+ Dynamic clipping & 3.571 & +0.181 \\
+ MBS & 3.386 & -0.004 \\
+ Low-rank compensation ($r=32$) & 3.397 & +0.007 \\
+ M+Adam & 3.383 & -0.007 \\
+ De-oscillation & \textbf{3.382} & \textbf{-0.008} \\
    \bottomrule
  \end{tabular}
  \label{tab:ablation-gbs16}
\end{table}

\paragraph{Summary.} The recommended ALTO recipe for GPT-OSS-20B MXFP4
training is \textbf{1d2d + RHT + SR + de-oscillation}. At GBS=16, this
configuration approaches BF16 validation loss within 0.007 at 16k steps
under simulated MXFP4 execution, and reaches the 3.34 target with
only a +5.0\% step overhead over BF16, versus +20.0\% for the other FP4
recipes (Table~\ref{tab:steps-to-target-valloss}). Techniques that improve synthetic operator SNR (static clipping, low-rank compensation) do not translate to better convergence at this scale,
while de-oscillation provides a consistent late-training benefit
(\(\Delta = -0.008\)) with only additional optimizer-state memory.

\subsection{Technique Evaluation and Recipe Selection}\label{technique-evaluation-and-recipe-selection}

This section consolidates end-to-end findings for the recommended stack
(Sections~\ref{hybrid-1d2d-block-quantization}~--~\ref{weight-de-oscillation}) and the additional techniques (Sections~\ref{differential-gradient-estimation-dge}~--~\ref{low-rank-outlier-compensation}).

\paragraph{Stochastic rounding.} Despite lowering input-gradient SNR (Section~\ref{operator-level-numerical-accuracy}), SR remains in the recommended recipe: it removes systematic gradient-quantization bias, and the recommended MXFP4 stack without de-oscillation already tracks BF16 closely (Figure~\ref{fig:valloss16}).

\paragraph{Weight de-oscillation vs.~low-rank compensation.} These methods
impose qualitatively different costs. Low-rank compensation adds
per-step compute that scales with rank \(r\) (Section~\ref{low-rank-outlier-compensation}).
De-oscillation adds mainly optimizer-state memory (\(d_w\), \(d_Q\), and
a weight snapshot) with periodic in-place snaps and no extra GEMMs. On
AMD MI300/MI355, GPT-OSS-20B training leaves sufficient memory headroom
for the auxiliary de-oscillation state without reducing batch size or
sequence length.

In our ablations (Table~\ref{tab:ablation-gbs16}), de-oscillation delivers the largest improvement, whereas low-rank compensation at $r=32$ is slightly harmful. We therefore adopt de-oscillation as the late-training stabilizer.

\paragraph{Weight de-oscillation vs. M+Adam.} The two approaches are comparable in convergence speed. However, the MLPerf Small MoE Training closed division requires the optimizer to be exactly AdamW; M+Adam does not satisfy this constraint. We therefore prefer weight de-oscillation for closed-division submission, where it can be applied as a post-step hook on standard AdamW without modifying the optimizer itself. For open-division submission, where optimizer choice is unrestricted, we prefer M+Adam as a cleaner, optimizer-native alternative that achieves equivalent convergence without auxiliary de-oscillation state.

\paragraph{Differential gradient estimation.} Operator-level tests show
only a marginal SNR improvement on the weight gradient, but validation
loss degrades severely (Table~\ref{tab:ablation-gbs16}), confirming that
the corrected DGE formula (Section~\ref{differential-gradient-estimation-dge}) does not help at GPT-OSS-20B scale.

\paragraph{Outlier clipping.} Both modes degrade operator-level SNR (Section~\ref{operator-level-numerical-accuracy}). Static clipping is roughly neutral on validation loss, while dynamic clipping causes a large regression Table~\ref{tab:ablation-gbs16}.

\paragraph{Macro-block scaling.} MBS yields the largest operator-level SNR gains of any technique (Section~\ref{operator-level-numerical-accuracy}) and translates into a small validation-loss improvement as well (Table~\ref{tab:ablation-gbs16}).

\paragraph{Operator--end-to-end gap.} Higher synthetic operator SNR is generally expected to benefit end-to-end training, but the correspondence is not guaranteed. In our experiments this link breaks down: \textbf{MBS} and \textbf{static clipping} raise operator SNR yet only marginally affect validation loss at GBS=16, while \textbf{DGE} and \textbf{dynamic clipping} are actively harmful. This indicates that operator-level SNR is a useful but imperfect proxy for end-to-end quality at GPT-OSS-20B scale. \textbf{De-oscillation} provides the most reliable late-training benefit among the techniques evaluated.

\paragraph{Preferred recipe.} Weighing final validation loss (Figure~\ref{fig:valloss16}), steps-to-target (Table~\ref{tab:steps-to-target-valloss}), and compute/memory cost, the preferred quantization stack for GPT-OSS-20B FP4 training is
\textbf{MXFP4 with 1d2d hybrid block quantization + RHT + SR}. For the training stabilizer, we recommend \textbf{weight de-oscillation} in MLPerf closed-division settings and \textbf{M+Adam} in open-division settings; both reach the 3.34 quality target in 16,128 steps at GBS=16. The remaining techniques of Sections~\ref{differential-gradient-estimation-dge}~--~\ref{low-rank-outlier-compensation} are not included in the preferred recipe.

\section{Limitations}\label{limitations}

\paragraph{Simulated execution and speed.} Current implementation simulate MXFP4 GEMM through per-operand quantize-dequantize round trips, since our MI300X hardware lacks native MXFP4 GEMM support. This models the numerical error budget of FP4 training faithfully, but not its memory-bandwidth or compute savings, and the training path itself has not been optimized for throughput (separate, unfused kernel launches for quantization, RHT, and GEMM). So the accuracy results here speak to algorithmic robustness, not the behavior of a fused production stack, and we do not report a wall-clock speedup over BF16.

\paragraph{Wall-clock performance.} The training path has not been
optimized for throughput. Each linear and grouped-GEMM layer incurs
separate kernel launches for quantization/dequantization, RHT and GEMM.
Auxiliary operations such as RHT are not fused with quantization or GEMM. Among the
techniques evaluated here, low-rank outlier compensation
is the primary source of auxiliary compute cost, adding per-layer GEMMs
on every training step; weight de-oscillation adds mainly memory cost
through extra optimizer-state tensors, a modest increase that fits
within available memory headroom on AMD MI300/MI355 hardware under our
GPT-OSS-20B configuration. We therefore do not report training-time
speedup over BF16: any observed wall-clock ratio would reflect prototype
overhead rather than the theoretical advantage of MXFP4 arithmetic. A
production implementation would require operator fusion, persistent
MXFP4 buffer management, and elimination of redundant
quantization/dequantization.

\paragraph{Scope of evaluation.} Results are reported for a single model
family (GPT-OSS-20B MoE) on a C4 subset, on AMD MI300 accelerators.
Generalization to dense architectures, other FP4 formats, larger global batch sizes and longer
training horizons remains to be established. DGE, clipping, and
macro-block scaling~\cite{chhugani2026unveiling} may warrant re-evaluation
under different hyperparameter regimes; low-rank compensation requires further study across ranks and on grouped GEMM.

\section{Conclusion}\label{conclusion}

We have described an ALTO~\cite{alto} recipe for training GPT-OSS-20B under MXFP4 on
the MLPerf Small MoE training benchmark. \textbf{2D block quantization, RHT, and
SR} are adopted from Abecassis, Felix et al.~\cite{abecassis2025pretraining} and form a strong baseline.
At GBS=16 and 16,128 steps, this stack reaches validation loss 3.3418
versus 3.3283 for BF16 (+0.014). \textbf{Weight de-oscillation}, adapted
from OsciReset in TetraJet-v2~\cite{chen2025tetrajet}, closes half of the
remaining gap, reaching a final loss \textbf{3.3350} (+0.007 vs.~BF16) with only
modest additional optimizer-state memory and no extra GEMMs. It also
sharply reduces the steps-to-target overhead: reaching the 3.34 quality
target costs only \textbf{+5.0\%} more steps than BF16 at GBS=16 (down
from \textbf{+20.0\%} for plain MXFP4).

Our ablations reveal that higher synthetic operator SNR does not always translate into better end-to-end training: \textbf{MBS} and \textbf{static clipping} raise operator SNR but do not improve validation loss at GBS=16, while \textbf{DGE} and \textbf{dynamic clipping} are actively harmful at scale. \textbf{Low-rank compensation} at $r=32$ is neutral to slightly harmful at GBS=16 and trails \textbf{de-oscillation} ($\Delta = -0.008$) in both accuracy and compute efficiency. De-oscillation recovers most of the loss gap, supporting its use as the late-training stabilizer in the recommended recipe: \textbf{1d2d + RHT + SR + de-oscillation}.

Future work will fuse quantization and transforms into a single kernel
pipeline, enable FP4 dispatch in expert parallelism, and pursue audited
MLPerf Training submission on native FP4 hardware.

\bibliographystyle{unsrt}  
\bibliography{references}

\end{document}